\PassOptionsToPackage{T2A}{fontenc}
\documentclass[onecolumn]{cohere}

\usepackage{lipsum}
\usepackage{pdfpages}
\usepackage{colortbl}
\usepackage{tabulary}
\usepackage{longtable}
\usepackage{array}
\usepackage{placeins}
\usepackage{tablefootnote}
\usepackage{tcolorbox}
\usepackage{wrapfig}
\usepackage{breqn} 

\usepackage{pifont}
\definecolor{deeppurple}{HTML}{9e02f7}
\definecolor{forestgreen}{HTML}{2e7d43}
\usepackage{multirow}
\usepackage{tabularx}

\usepackage[utf8]{inputenc}

\usepackage{arydshln}

\usepackage{xspace} 
\usepackage{makecell} 
\usepackage{multirow}

\usepackage[framemethod=TikZ]{mdframed}
\usepackage{tcolorbox}
\usepackage{multicol}

\newtcolorbox{mybox}[2][]{
  colback=white, 
  colframe=lightblue,
  fonttitle=\bfseries,
  coltitle=black,  
  title=#2, 
  #1
}

\definecolor{ayad}{RGB}{148, 156, 229} 
\definecolor{ayadsymbol}{RGB}{76, 110, 230} 
\definecolor{lightblue}{RGB}{211, 227, 252} 

\definecolor{bgblue}{RGB}{247, 250, 255} 
\newcommand*\colourcheck[1]{%
  \expandafter\newcommand\csname #1check\endcsname{\textcolor{#1}{\ding{52}}}%
}
\newcommand*\colourcross[1]{%
  \expandafter\newcommand\csname #1cross\endcsname{\textcolor{#1}{\ding{55}}}%
}
\DeclareSymbolFont{extraup}{U}{zavm}{m}{n}
\DeclareMathSymbol{\vardiamond}{\mathalpha}{extraup}{87}
\definecolor{ayadsymbol}{RGB}{76, 110, 230} 

\colourcheck{blue}
\colourcheck{green}
\colourcheck{red}

\colourcross{blue}
\colourcross{green}
\colourcross{red}
\usepackage{nicematrix}
\usepackage{comment}
\usepackage{amssymb}

\setcitestyle{number,square}

\title{\includegraphics[scale=0.2]{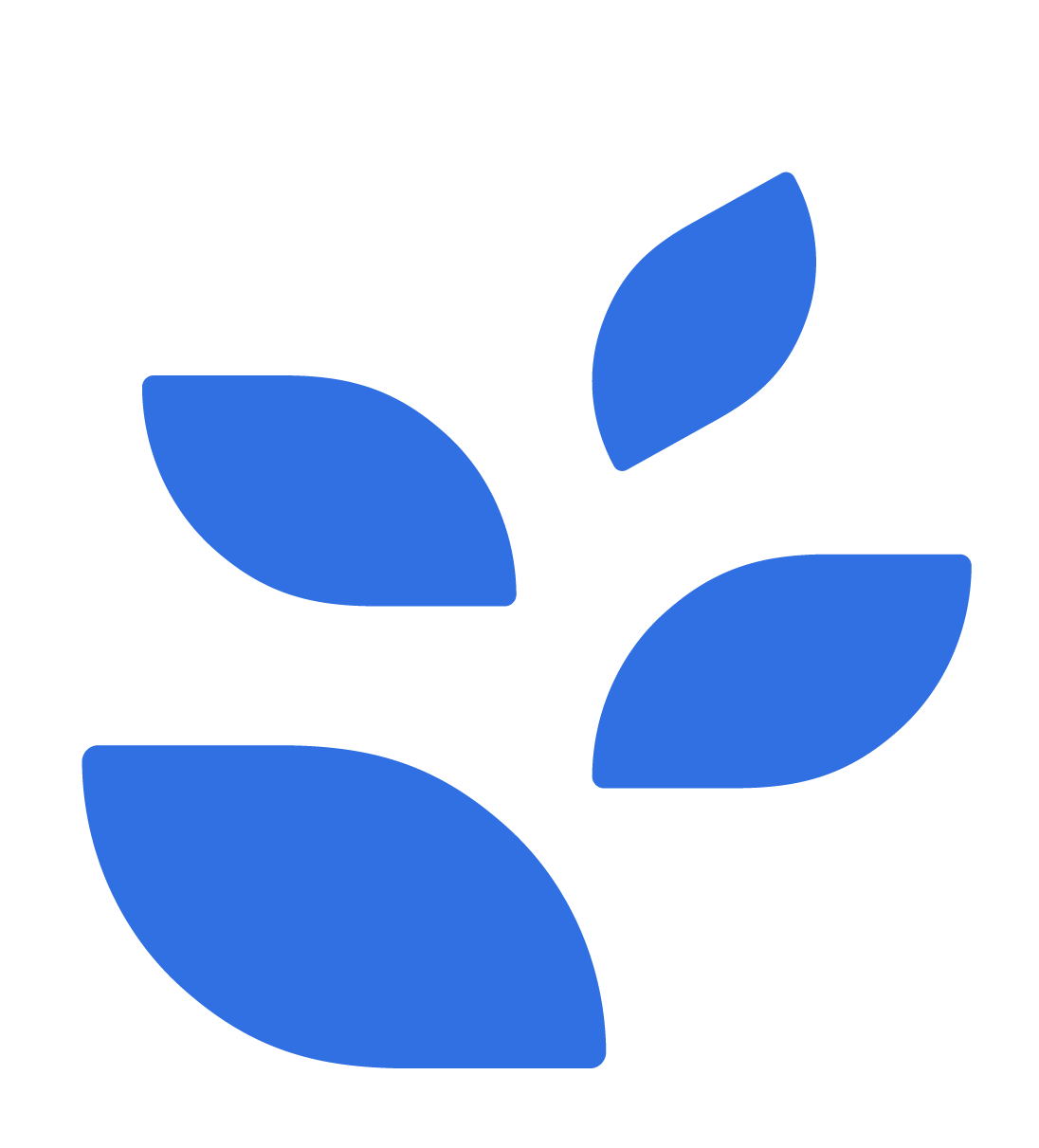}Aya Expanse: Combining Research Breakthroughs for a New Multilingual Frontier}

\renewcommand{\fasymbol}{\textcolor{Cerulean}{$\vardiamond$}}
\renewcommand{\fasymbola}{\textsuperscript{\textcolor{NavyBlue}{$\bigstar$}}}

\multiauthors
\author{
    name={John Dang\fa},
    affiliation={1},
}
\author{
    name={Shivalika Singh\fa},
    affiliation={1},
}
\author{
    name={Daniel D'souza\fa},
    affiliation={1},
}
\author{
    name={Arash Ahmadian\fa},
    affiliation={1},
}
\author{
    name={Alejandro Salamanca},
    affiliation={1},
}
\author{
    name={Madeline Smith},
    affiliation={1},
}
\author{
    name={Aidan Peppin},
    affiliation={1},
}
\author{
    name={Sungjin Hong},
    affiliation={2},
}
\author{
    name={Manoj Govindassamy},
    affiliation={2},
}
\author{
    name={Terrence Zhao},
    affiliation={2},
}
\author{
    name={Sandra Kublik},
    affiliation={2},
}
\author{
    name={Meor Amer},
    affiliation={2},
}
\author{
    name={Viraat Aryabumi},
    affiliation={2},
}
\author{
    name={Jon Ander Campos},
    affiliation={2},
}
\author{
    name={Yi-Chern Tan},
    affiliation={2},
}
\author{
    name={Tom Kocmi},
    affiliation={2},
}
\author{
    name={Florian Strub},
    affiliation={2},
}
\author{
    name={Nathan Grinsztajn},
    affiliation={2},
}
\author{
    name={Yannis Flet-Berliac},
    affiliation={2},
}
\author{
    name={Acyr Locatelli},
    affiliation={2},
}
\author{
    name={Hangyu Lin},
    affiliation={2},
}
\author{
    name={Dwarak Talupuru},
    affiliation={2},
}
\author{
    name={Bharat Venkitesh},
    affiliation={2},
}
\author{
    name={David Cairuz},
    affiliation={2},
}
\author{
    name={Bowen Yang},
    affiliation={2},
}
\author{
    name={Tim Chung},
    affiliation={2},
}
\author{
    name={Wei-Yin Ko},
    affiliation={2},
}
\author{
    name={Sylvie Shang Shi},
    affiliation={2},
}
\author{
    name={Amir Shukayev},
    affiliation={2},
}
\author{
    name={Sammie Bae},
    affiliation={2},
}
\author{
    name={Aleksandra Piktus},
    affiliation={2},
}
\author{
    name={Roman Castagné},
    affiliation={2},
}
\author{
    name={Felipe Cruz-Salinas},
    affiliation={2},
}
\author{
    name={Eddie	Kim},
    affiliation={2},
}
\author{
    name={Lucas Crawhall-Stein},
    affiliation={2},
}
\author{
    name={Adrien Morisot},
    affiliation={2},
}
\author{
    name={Sudip Roy},
    affiliation={2},
}
\author{
    name={Phil Blunsom},
    affiliation={2},
}
\author{
    name={Ivan Zhang},
    affiliation={2},
}
\author{
    name={Aidan Gomez},
    affiliation={2},
}
\author{
    name={Nick Frosst},
    affiliation={1,2},
}

\author{
    name={Marzieh Fadaee\fasymbola},
    affiliation={1},
}
\author{
    name={Beyza Ermis\fasymbola},
    affiliation={1},
}

\author{
    name={Ahmet Üstün\fasymbola},
    affiliation={1},
}
\author{
    name={Sara Hooker\fasymbola},
    affiliation={1},
}

\affiliations{
    \item[1] Cohere For AI 
    \item[2] Cohere
}

\corresponding[*]{Ahmet Üstün \texttt{<ahmet@cohere.com>}, Sara Hooker \texttt{<sarahooker@cohere.com>}}

\date{\today}

\abstract{
We introduce the Aya Expanse model family, a new generation of 8B and 32B parameter multilingual language models, aiming to address the critical challenge of developing highly performant multilingual models that match or surpass the capabilities of monolingual models. By leveraging several years of research at Cohere For AI and Cohere, including advancements in data arbitrage, multilingual preference training, and model merging, Aya Expanse sets a new state-of-the-art in multilingual performance. Our evaluations on the Arena-Hard-Auto dataset, translated into 23 languages, demonstrate that Aya Expanse 8B and 32B outperform leading open-weight models in their respective parameter classes, including Gemma 2, Qwen 2.5, and Llama 3.1, achieving up to a 76.6\% win-rate. Notably, Aya Expanse 32B outperforms Llama 3.1 70B, a model with twice as many parameters, achieving a 54.0\% win-rate. In this short technical report, we present extended evaluation results for the Aya Expanse model family and release their open-weights, together with a new multilingual evaluation dataset m-ArenaHard.     

\vspace{0.2cm}
\textbf{Aya Expanse 8B}: \url{https://hf.co/CohereForAI/aya-expanse-8b}

\textbf{Aya Expanse 32B}: \url{https://hf.co/CohereForAI/aya-expanse-32b}

\vspace{0.2cm}
\textbf{m-ArenaHard Dataset}: \url{https://hf.co/datasets/CohereForAI/m-ArenaHard}
}

\renewcommand{\FONTauthors}{\bfseries\Large}

\begin{document}

\section{Introduction}

We introduce the Aya Expanse, a family of multilingual instruction-tuned language models that support 23 languages, built upon the recent Cohere's Command series\footnote{\url{https://cohere.com/blog/command-series-0824}}. Developed by Cohere For AI and Cohere, the Aya Expanse models are an open-weights release of both 8-billion and 32-billion parameters that address the significant challenge of developing high-performance multilingual models that can rival their monolingual counterparts. 

Despite notable advancements in large language models~\citep{openai2023GPT4,team2023gemini,team2024gemma,dubey2024llama,yang2024qwen2}, there remains a stark gap in the performance of models across multiple languages. Models achieve superior performance on languages that they are trained on \citep{kunchukuttan-etal-2021-large}, however, this leads to biases against languages unseen during training \citep{schwartz2022towards, Kotek2023GenderBA, Khandelwal2023CasteistBN, vashishtha2023evaluating,khondaker2023gptaraeval}, and critical safety and security flaws for all users \citep{yong2023lowresource, nasr2023scalable, Li2023PrivacyIL, Lukas2023AnalyzingLO, deng2023multilingual}.
Furthermore, languages that the models are not optimized for, require more tokens due to poor tokenization, leading to higher latency and cost which limits the use of this technology even further \citep{held2023material, durmus2023measuring,nicholas2023lost,ojo2023good}.  

The Aya Expanse model family seeks to mitigate this gap through a suite of innovative methodologies, including multilingual data arbitrage~\citep{odumakinde2024multilingualarbitrageoptimizingdata}, multilingual preference optimization and safety tuning~\citep{dang2024rlhf,aakanksha2024multilingual}, and model merging~\citep{aakanksha2024mix}. In this work, we provide extended evaluations for our work closing the language gap through Aya Expanse. As the main evaluation dataset, we use the Arena-Hard-Auto dataset \citep{li2024crowdsourced} and translate it into 23 languages to enable extensive testing across varied linguistic contexts. In addition to multilingual preference evaluation on m-ArenaHard and Dolly \citep{ayadata2024} datasets, we report a series of multilingual academic benchmarks following to \citet{ustun2024aya} and \citet{aryabumi2024aya}. 

\begin{figure}[t]
     \centering
     \begin{subfigure}[b]{0.49\textwidth}
          \centering
          \includegraphics[width=\textwidth]{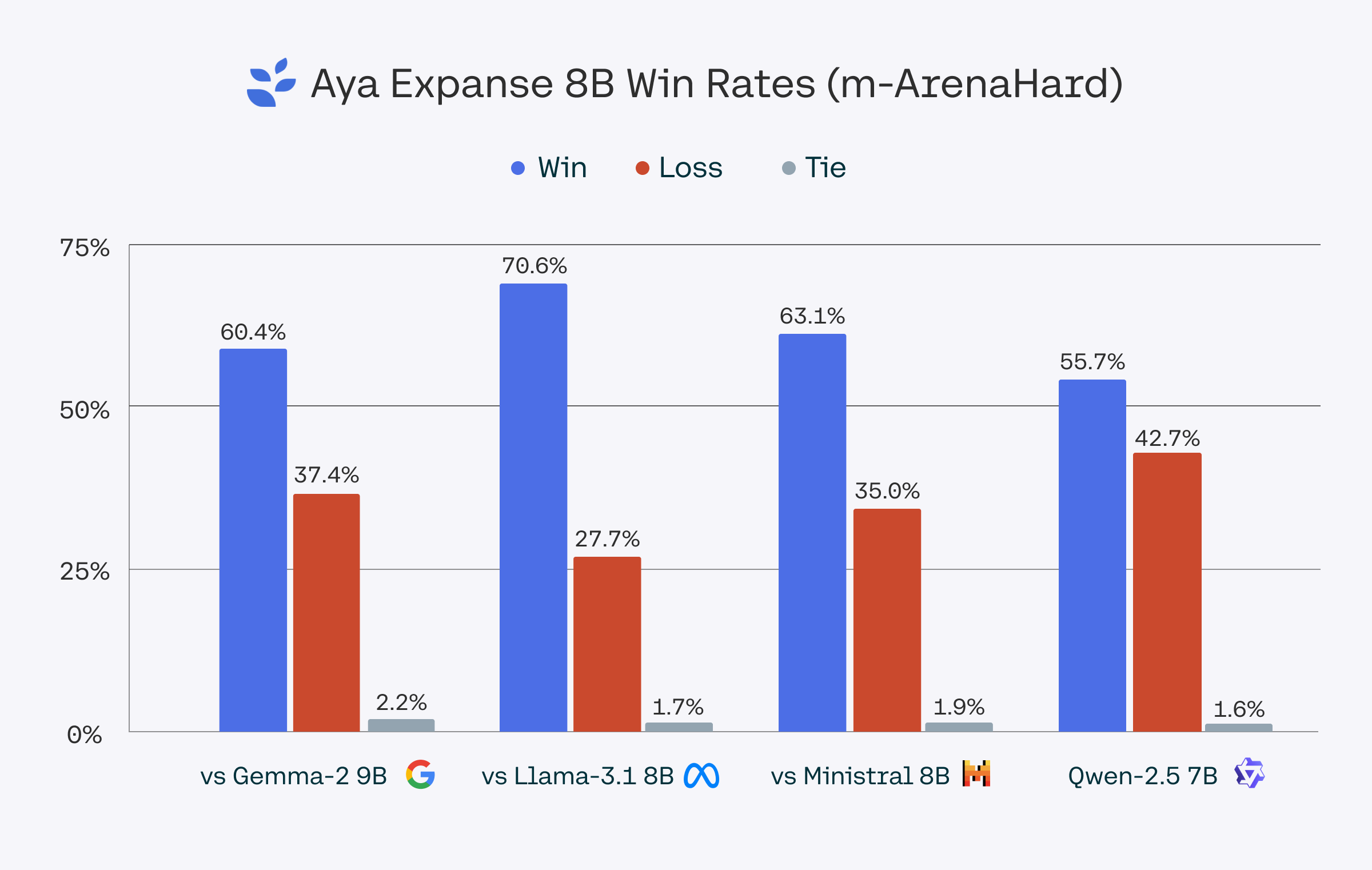}
      \end{subfigure}
      \hspace{0.1cm}
     \begin{subfigure}[b]{0.49\textwidth}
          \centering
          \includegraphics[width=\textwidth]{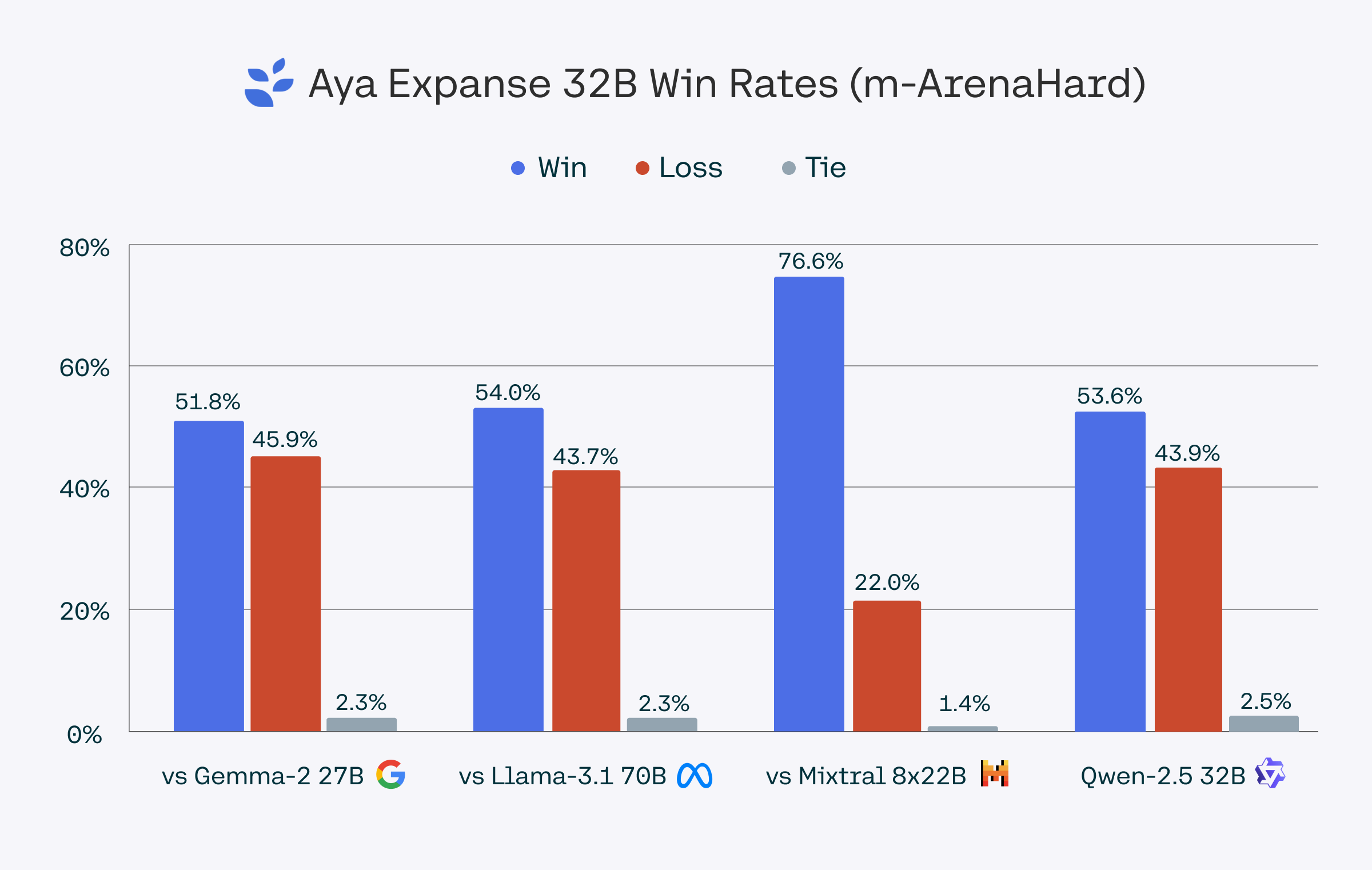}
      \end{subfigure}
      \caption{\textbf{Pairwise win-rates on m-ArenaHard averaged across 23 languages}: We compare Aya Expanse 8B (left) with Gemma 2 9B, Llama 3.1 8B, Ministral 8B and Qwen 2.5 7B. Aya Expanse 32B (right) is compared with Gemma 2 27B, Qwen 2.5 32B, Mixtral 8x22B, and Llama 3.1 70B. We used the instruction-tuned version of all models.}
      \label{fig:m-arenahard}
 \end{figure}

In direct head-to-head win rate evaluations across 23 languages, Aya Expanse 8B and 32B models outperform the similar-sized leading open-weight models by up to 76.6\% win-rates against Gemma 2 9B and 27B \citep{team2024gemma}, Qwen 2.5 7B and 32B \citep{qwen2.5}, Mistral models \citep[8B, 8x22B;][]{ministral,mixtral}, and Llama 3.1 8B and 70B \citep{dubey2024llama} -- even though the Llama's largest variant includes over twice as many parameters than our 32B. Additionaly, in multilingual academic benchmarks, Aya Expanse models achieve a significant advancement compared to their predecessor Aya 23 models \citep{aryabumi2024aya} up to 16\% increase in language understanding and 1.8x increase in mathematical reasoning, enabling competitive performance with the other open-weight models.    

We release open-weights for Aya Expanse 8B and 32B together with m-ArenaHard, the translated Arena-Hard-Auto dataset. By making these resources available, we aim to accelerate progress in the field of multilingual AI and support the development of more inclusive language technologies.

\section{Post-Training Recipe} \label{sec:post-training}

Our post-training recipe for Aya Expanse leverages critical details from research papers we have released. We briefly describe these core drivers of performance below, and refer the reader to the original manuscripts for a more detailed understanding.

\subsection{Synthetic Data Generation through Data Arbitrage 
}

The use of synthetic data – data generated by an expert or ``teacher'' model to train another model – has become increasingly central to the development of LLMs, particularly as model training has exhausted current data sources. However, for multilingual data, especially with low-resource languages, there are few good examples of teacher models, creating an extra added challenge to leveraging synthetic data. Furthermore, recent research has suggested that an over-reliance on synthetic data leads to model collapse \citep{792c074246bd4d3db5d46a5bdfee2646}.

In our recent work \citep{odumakinde2024multilingualarbitrageoptimizingdata}, we demonstrate that these limitations can be addressed through “data arbitrage” – strategically sampling from a pool of teacher models. This approach has important implications as it challenges the traditional reliance on a single-teacher model for generating synthetic data. Instead, data arbitrage leverages performance variations among a pool of models. Although this technique is applicable to any domain, it is particularly suited to the multilingual setting, where the absence of a universally effective teacher that excels across all languages presents significant challenges. In the creation of high-quality synthetic multilingual datasets, multilingual arbitrage proves valuable by utilizing a diverse pool of models to strategically sample different parts of the data distribution for improved multilingual generations.

We first train a model pool for groups of languages and employ an \textbf{Arbiter} to evaluate and select the optimal generation. The Arbiter here is an internal Reward Model to score the model generations. In Reward-Based Routing, for each prompt in a given language, we generate completions from all models in the pool and score them using the reward model. The completion with the highest score is chosen as the final completion for that prompt. As shown in Figure \ref{fig:vs-aya-23} (Section \ref{sec:improve}), Aya Expanse 8B model, even at the SFT stage trained with Multilingual Arbitrage, had over 9.1\% improvement in win-rate measured against Gemma 2 9B compared to the previous Aya 23 model \citep{aryabumi2024aya}, demonstrating the effectiveness of this approach in leveraging diverse model strengths across languages.

\subsection{Iterative Multilingual Preference Training}

Following supervised fine-tuning, alignment to human preferences is a key step for training today’s state-of-the-art LLMs. Although heavily adopted, it is known that preference training is already challenging in a monolingual setting \citep{casper2023open,ahmadian-etal-2024-back}. Maximizing gains from preference training in a multilingual setting introduces even more challenges. The vast majority of existing preference datasets are exclusively English and the few existing multilingual preference datasets are often of low-quality. Moreover, modeling many diverse languages simultaneously is known to be a difficult optimization problem where naively optimizing for performance in some languages often leads to regressions in performance in other languages.

In several recent works \citep{dang2024rlhf,aakanksha2024multilingual} we explore extending preference training to a massively multilingual setting. In our recent work \citep{dang2024rlhf}, we leverage a novel synthetic data generation technique to construct high-quality multilingual preference data pairs by contrasting in-language completions from a highly performant multilingual LLM with lower quality completions translated from English which were generated by a weaker model. This steers our model away from generating low-quality multilingual completions which often contain undesirable artifacts, such as those introduced by poor translation. We show that this method unlocks substantial gains in performance across all languages and often also results in gains for languages not included in the preference training data.

While \citet{dang2024rlhf} also show that preference training with online data outperforms its offline variant, during training of Aya Expanse, we found that the combination of first preference-training with offline data followed by preference-training with online data to be better than either online or offline training alone. In the first preference training stage, we train on data curated by taking the highest and lowest reward responses from the Arbitrage stage as the chosen and rejected completions, which makes the first stage of DPO training \textit{offline}.

After offline preference training, we run \textit{online} iterative DPO, where we sample multiple online generations for each prompt from the model trained during the last iteration, rank these generations with an internal Reward Model, and then further train on these preference pairs. For both models, we repeat this process for 3 iterations as we found that going beyond 3 iterations led to minimal gains at the cost of additional re-tuning parameters like regularization coefficient ($\beta$) and sometimes introduced reward hacking behavior. Overall, for Aya Expanse 8B, the combination of offline and online preference training on top of the model trained with arbitrage, led to 7.1\% additional gains in win rate against Gemma 2 9B (Figure \ref{fig:vs-aya-23}, Section \ref{sec:improve}).

\subsection{Model Merging}

A reappearing problem throughout any post-training (and pre-training) pipeline, whether it consists of a single stage such as SFT, or a more complex multi-stage optimization pipeline, such as our pipeline above, is choosing the right data mixtures for training. The intricacies of this process demand considerable effort in fine-tuning hyperparameters and data combinations. Merging multiple models is an alternative approach for enabling complex multi-tasking at a reduced aggregate computational cost. In Aya Expanse, we directly build on the findings of our recent research paper \citep{aakanksha2024mix} and apply merging in both the Arbitrage phase, and at each iteration of preference training.

When training multiple separate models with the goal of merging, it is important to maximize diversity between checkpoints. However, this should be balanced with ensuring that each individual model within the pool achieves high performance. To balance these objectives, we maximize diversity between checkpoints by training models for different language families. This takes advantage of cross-lingual transfer which often provides significant performance benefits while ensuring that linguistic differences provide sufficient differentiation between checkpoints.

Naively, one could split-train a model for each language and then merge, but this does not achieve the same benefits we observe from cross-lingual transfer. To improve robustness in merging, we include some shared languages across each cluster (English, Spanish, and French). In the final recipe, we used multiple stages of merging runs trained on different clusters of data, and checkpoints within the same run.

In addition to weighted linear averaging, we experiment with multiple merging techniques, namely SLERP \citep{10.1145/325334.325242}, TIES-merging \citep{yadav2023tiesmergingresolvinginterferencemerging}, and DARE-TIES \citep{yu2024languagemodelssupermario}. However, we found weighted averaging to be the most consistent method. As a result, we use weighted averaging throughout the pipeline. Interestingly, we observed significantly larger gains from merging at the 32B scale compared to the 8B scale – up to 3x. This is in line with recent work suggesting merging to be more effective at scale \citep{yadav2024mattersmodelmergingscale}.

\section{Model Architecture and Experimental Details}

The Aya Expanse model family is based on the recent Cohere Command series\footnote{\url{https://cohere.com/blog/command-series-0824}} which are pre-trained using a data mixture that includes texts from 23 languages. These 23 languages are: \textit{Arabic, Chinese (simplified \& traditional), Czech, Dutch, English, French, German, Greek, Hebrew, Hindi, Indonesian, Italian, Japanese, Korean, Persian, Polish, Portuguese, Romanian, Russian, Spanish, Turkish, Ukrainian} and \textit{Vietnamese}.

Regarding the models used as the base for Aya Expanse, the 8B model is trained with a maximum context length of 8K, while the 32B model is trained with a maximum context length of 128K. Similar to previous Aya 23 models \citep{aryabumi2024aya}, both 8B and 32B models use SwiGLU activation function \citep{gluvariants}, RoPE positional embeddings \citep{rope} and grouped-query attention \citep{gqa}. For the instruction fine-tuning, we format prompt-completion pairs by using the chat template that is described in \citet{aryabumi2024aya} where roles (\texttt{user}, \texttt{chatbot}), and chat turns are demonstrated with special tokens.  

\section{Multilingual Evaluation}
\begin{table}
    \centering
    \resizebox{\textwidth}{!}{
    \begin{NiceTabular}{llccccc>{\columncolor{forestgreen!60}}c>{\columncolor{forestgreen!40}}c>{\columncolor{forestgreen!20}}c}[colortbl-like]
        \toprule
        Task & Dataset & \multicolumn{2} {c} {Metric} & Languages \\
        \midrule
        \noalign{\smallskip}
        \textsc{\textbf{Discriminative Tasks}} \\
        \noalign{\smallskip} 
        Coreference Resolution & XWinograd~\citep{muennighoff2022crosslingual} & 0-shot & Acc.  & 6  \\
        \multirow{2}{*}{Sentence Completion} & XCOPA~\citep{ponti2020xcopa} &  0-shot & Acc.  & 11  \\
        & XStoryCloze~\citep{lin2021fewshot} & 0-shot & Acc.  & 10  \\
        \noalign{\smallskip} 
        \hdashline 
        \noalign{\smallskip}
        \multirow{2}{*}{Language Understanding} & Global-MMLU~\citep{singh2024globalmmluunderstandingaddressing} & 5-shot & Acc.  & 23  \\
        & INCLUDE~\citep{romanou2024include} & 0-shot & Acc.  & 44  \\
        \midrule
        \textsc{\textbf{Generative Tasks}} \\
        \noalign{\smallskip} 
        Translation & FLORES-200~\citep{goyal2021flores101,nllb2022} & 0-shot & chrF\texttt{++}, xCOMET  & 23 \\
        \noalign{\smallskip} 
        \hdashline 
        \noalign{\smallskip}
        Mathematical Reasoning & MGSM~\citep{shi2023language-mgsm} & 5-shot & Acc.  & 7 \\
        \noalign{\smallskip} 
        \hdashline 
        \noalign{\smallskip}
        \multirow{2}{*}{Open-Ended Generation} & Multilingual ArenaHard & 0-shot & win-rate  & 23  \\
        & Dolly Human-edited \& Machine-translated~\citep{ayadata2024}  & 0-shot & win-rate  & 23  \\
        \bottomrule
    \end{NiceTabular}
    }
    \caption{\textbf{Comprehensive evaluation suite including 8 datasets, 6 task categories for up to 23 languages}: Datasets considered for evaluation following \citep{ustun2024aya,aryabumi2024aya}. Additionally, we included multilingual ArenaHard which is a translated version of the original Arena-Hard-Auto dataset \citep{li2024crowdsourced}. We limit the evaluation languages only to the ones that are included in 23 languages, except for XWinograd, XCOPA, XStoryCloze and INCLUDE where we use all the available languages. 
    }
    \label{tab:benchmarks}
\end{table}
To measure the multilingual performance of Aya Expanse models and compare them to other models, we follow the comprehensive evaluation framework introduced by \citet{ustun2024aya}. We follow the same methodology and implementation used by \citet{aryabumi2024aya} and also include a new dataset, m-ArenaHard, to evaluate open-ended generation performance in addition to Dolly \citep{ayadata2024}. 
Details of evaluation categories are as follows:   

\begin{enumerate}
\item \textbf{Preference evaluation}: We measure the open-ended generation capabilities of the models through preference evaluation using LLM-as-a-Judge following the previous work \citep{rafailov2023direct,dubois2023alpacafarm,kim2023prometheus,ustun2024aya}. In particular, \citet{ustun2024aya} shows a high correlation between LLML-judge and human annotators for multilingual evaluation. As LLM-judge, we experimented with both GPT-4o-mini and GPT-4o. As GPT-4o-mini results significantly differ from GPT-4o where GPT-4o-mini as a judge was biased towards Aya Expanse, we opt for GPT-4o to report final results.\footnote{We used \texttt{gpt-4o-2024-08-06} as our judge model. Details: \url{https://platform.openai.com/docs/models/gpt-4o}}. We compare models for all 23 languages that Aya Expanse covers using two different datasets:  
\begin{enumerate}
    \item \textbf{Multilingual ArenaHard}: 
    We translate the original 500 English LMARENA (formerly LMSYS) Arena-Hard-Auto prompts into the other 22 languages supported by Aya Expanse using Google Translate API\footnote{Google Translate API v3 docs: \url{https://cloud.google.com/translate/docs/advanced/translating-text-v3}}. In contrast to the original Arena-Hard-Auto benchmark \citep{li2024crowdsourced} which measures win-rate against baseline completions from GPT-4-0314 with GPT-4-Turbo as a judge, we measure win-rates against the other models' generations directly using GPT-4o as a judge to get a more direct comparison with how well Aya Expanse performs across 23 languages with a newer, higher-quality judge model. 
    \item \textbf{Dolly Evaluation Sets}: We use \textbf{dolly-machine-translated} test set \citep{ayadata2024} which is a held-out test set of 200 instances from the Dolly-15k dataset \citep{DatabricksBlog2023DollyV2} 
translated into 101 languages. This test set was curated by multiple annotators to avoid the inclusion of any culturally specific or geographic references, intending to minimize estimations of performance that require specific cultural or geographic knowledge. For 6 languages (\texttt{French}, \texttt{Spanish}, \texttt{Serbian}, \texttt{Russian}, \texttt{Arabic}, \texttt{Hindi}), we use \textbf{dolly-human-edited} test set~\citep{ayadata2024}, which are improved versions of the \textbf{dolly-machine-translated} where automatic translations are post-edited by professional compensated human annotators to correct any possible translation issues. 
\end{enumerate}

We use the same prompt for eliciting GPT-4o preferences as specified by \citet{ustun2024aya}. 

\item \textbf{Discriminative tasks}: Following  \citet{aryabumi2024aya}, we evaluate on XWinograd~\citep{muennighoff2022crosslingual}, XCOPA~\citep{ponti2020xcopa}, and XStoryCloze~\citep{lin2021fewshot}. We use zero-shot evaluation. Note that these evaluation tasks are completely unseen and there is no dataset in the training mixture from the same task categories. 

\item \textbf{General purpose language understanding}: We evaluate the general language understanding of Aya Expanse models using Global-MMLU~\citep{singh2024globalmmluunderstandingaddressing} in a 5-shot evaluation setting, ensuring the dataset is not seen during training.
Global-MMLU is an improved version of the original MMLU dataset~\citep{hendrycks2020measuring}, which consists of 13,062 questions across 57 tasks covering STEM, humanities, and social sciences. It expands evaluation coverage to 42 languages by translating the English dataset into 41 additional languages ranging from low- to high-resource. Initial translations were performed using Google Translate API and then refined through contributions from compensated professional and community annotators, ensuring high-quality translations and addressing cultural biases in the original dataset.
For this evaluation, we focus on the 23 languages supported by the Aya Expanse models.

Additionally, to measure improvement from Aya 23 models \citep{aryabumi2024aya} to Aya Expanse, we use INCLUDE \citep{romanou2024include} dataset that consists of 22,637 questions extracted from in-language local academic and professional exams in 44 languages.

\item \textbf{Multilingual mathematical reasoning}: We use Multilingual Grade School Math (MGSM) Benchmark \citep{shi2023language-mgsm} to evaluate multilingual mathematical reasoning. MGSM consists of 250 problems from the GSM8K benchmark \citep{cobbe2021training}, which are human-translated into 10 languages. We pick the subset of 7 MGSM languages (English, German, Spanish, French, Japanese, Russian, Chinese), which are covered by Aya Expanse models. We use questions with answers followed by CoT prompt (\texttt{5-shot}) in the same language (\texttt{native_cot}) and \texttt{strict-match} score as the evaluation metric following \citet{shi2023language-mgsm}.

\item \textbf{Machine Translation}: We evaluate model performance in machine translation on FLORES-200~\citep{nllb2022}. For FLORES, we use all 22 languages (X $\leftrightarrow$ English) based on the language coverage of Aya Expanse models. As the evaluation metrics, we report both chrF\texttt{++} \citep{popovic-2017-chrf}, and xCOMET-XL \citep{guerreiro2024xcomet}\footnote{\url{https://huggingface.co/Unbabel/XCOMET-XL}} scores as it is one of the best-performing metrics for the machine translation evaluation strongly outperforming chrF\texttt{++} in assessing translation capabilities \citep{freitag-etal-2024-llms}.


\end{enumerate}

\begin{figure}[t]
     \centering
     \begin{subfigure}[b]{0.49\textwidth}
          \centering
          \includegraphics[width=\textwidth]{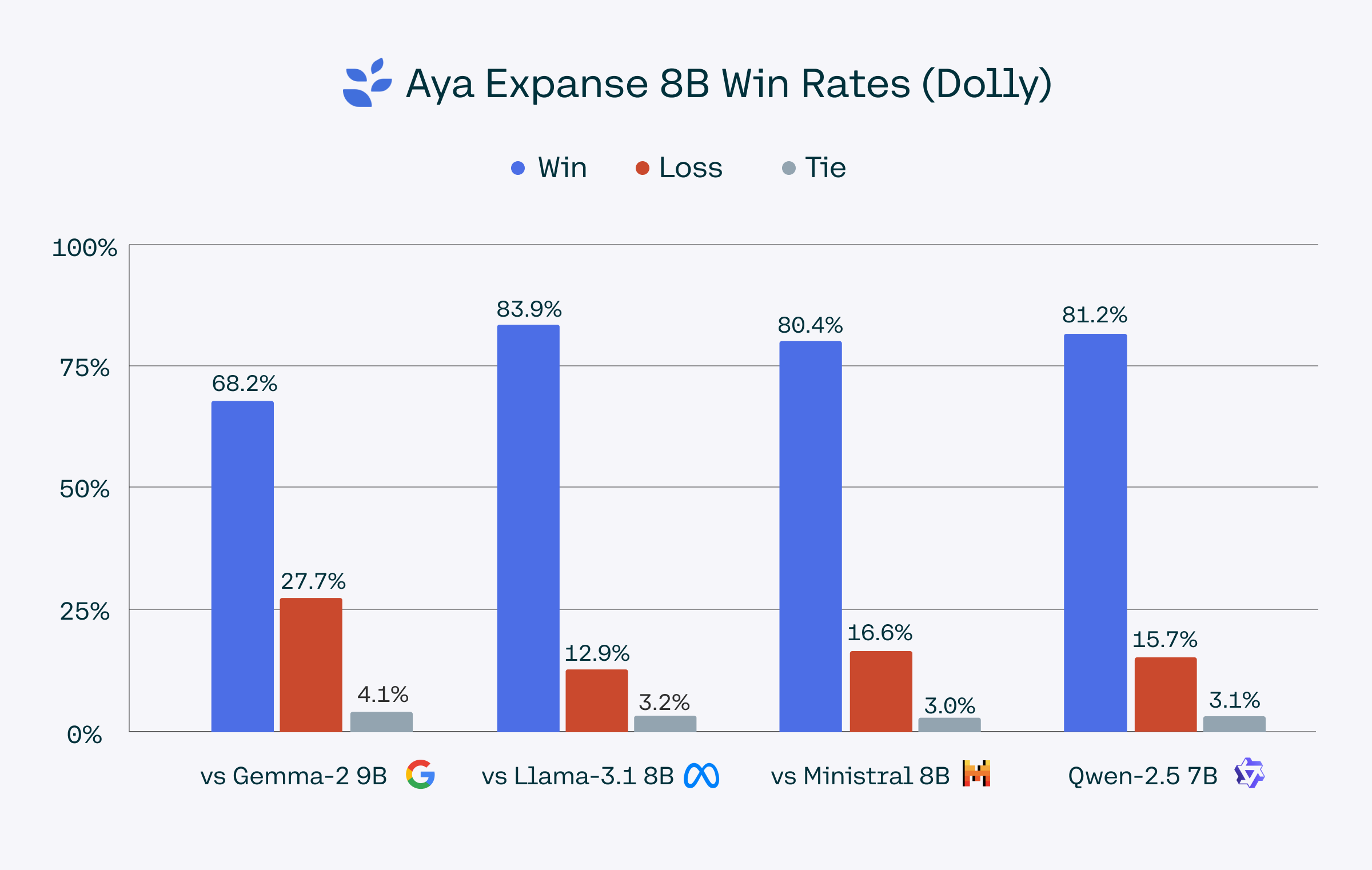}
      \end{subfigure}
      \hspace{0.1cm}
     \begin{subfigure}[b]{0.49\textwidth}
          \centering
          \includegraphics[width=\textwidth]{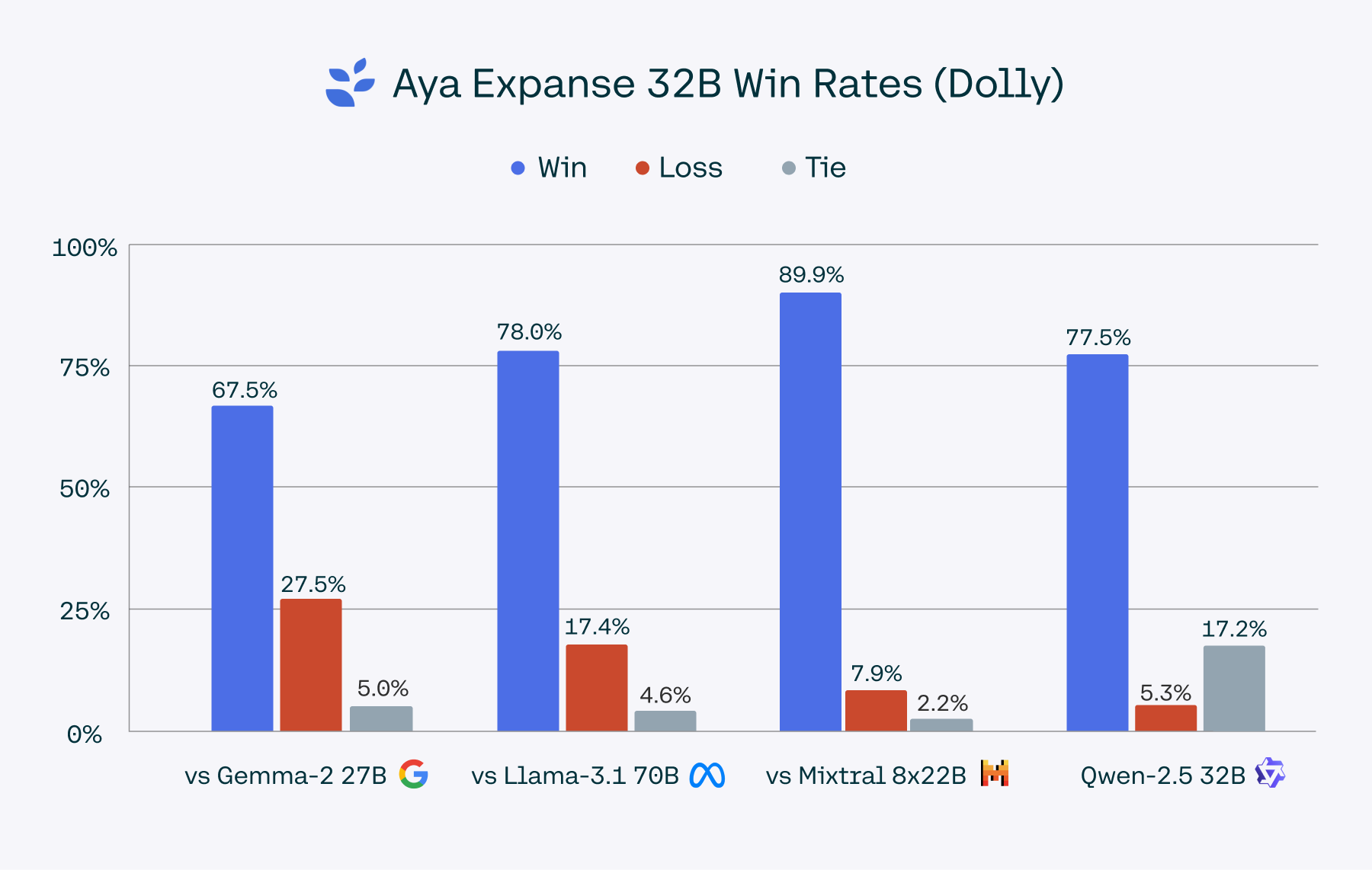}
      \end{subfigure}
      \caption{\textbf{Pairwise win-rates on Dolly evaluation set \citep{ayadata2024} averaged across 23 languages}: We compare Aya Expanse 8B (left) with Gemma 2 9B, Llama 3.1 8B, Ministral 8B and Qwen 2.5 7B. Aya Expanse 32B (right) is compared with Gemma 2 27B, Qwen 2.5 32B, Mixtral 8x22B and Llama 3.1 70B. We used the instruct fine-tuned (via SFT and RLHF) version of all models.}
      \label{fig:dolly}
 \end{figure}

\subsection{Model Comparisons}
We benchmark Aya Expanse models against leading open-weight models in both 8B and 32B parameter classes. Additionally, for Aya Expanse 32B, we also compared it with Mixtral 8x22B and Llama 3.1 70B even though these models include many more parameters. We briefly describe these models below.

\begin{enumerate}
    \item\textbf{Aya 23 (8B and 35B)}~\citep{aryabumi2024aya}: 8B and 35B parameter variants of our previous generation Aya 23 models. These models are optimized for 23 languages same as Aya Expanse models through supervised instruction fine-tuning.
    \item\textbf{Llama-3.1-Instruct (8B and 70B)}~\citep{dubey2024llama}: 8B and 70B parameter variants from the Llama 3.1 model released in July 2024. These models have also undergone alignment via Direct Preference Optimization (DPO; \citep{rafailov2023DPO}) and officially support English, German, French, Italian, Portuguese, Hindi, Spanish, and Thai.
    \item\textbf{Gemma-2-IT (9B and 27B)}~\citep{team2024gemma} 9B and 27B parameter variants from the Gemma 2 model released in June 2024. These models have undergone alignment via RLHF\citep{ouyang2022LLMRLHF} in addition to the SFT stage. Although they officially do not claim multilingual capability, they are often strong contenders in the setting.
    \item\textbf{Ministral-8B-Instruct-2410}~\citep{ministral}: A multilingual 8B instruction fine-tuned model released in October 2024 by Mistral AI. They officially support English, French, German, Spanish, Italian, Portuguese, Chinese, Japanese, Russian and Korean.
    \item\textbf{Mixtral-8x22B-Instruct-v0.1} \citep{mixtral}: A sparse Mixture-of-Experts(MoE) model released in August 2024. This model includes a total of 141B parameter where 39B active parameters is used per token. It supports English, French, Italian, German, and Spanish.
    \item\textbf{Qwen-2.5-Instruct (7B and 32B)}~\citep{qwen2.5}: 7B and 32B parameter variants from the Qwen 2.5 models released in September 2024. They have been instruction fine-tuned and support 29 languages including Chinese, English, French, Spanish, Portuguese, German, Italian, Russian, Japanese, Korean, Vietnamese, Thai, Arabic, and more.
\end{enumerate}

\begin{figure}[t]
     \centering
     \begin{subfigure}[b]{0.49\textwidth}
          \centering
          \includegraphics[width=\textwidth]{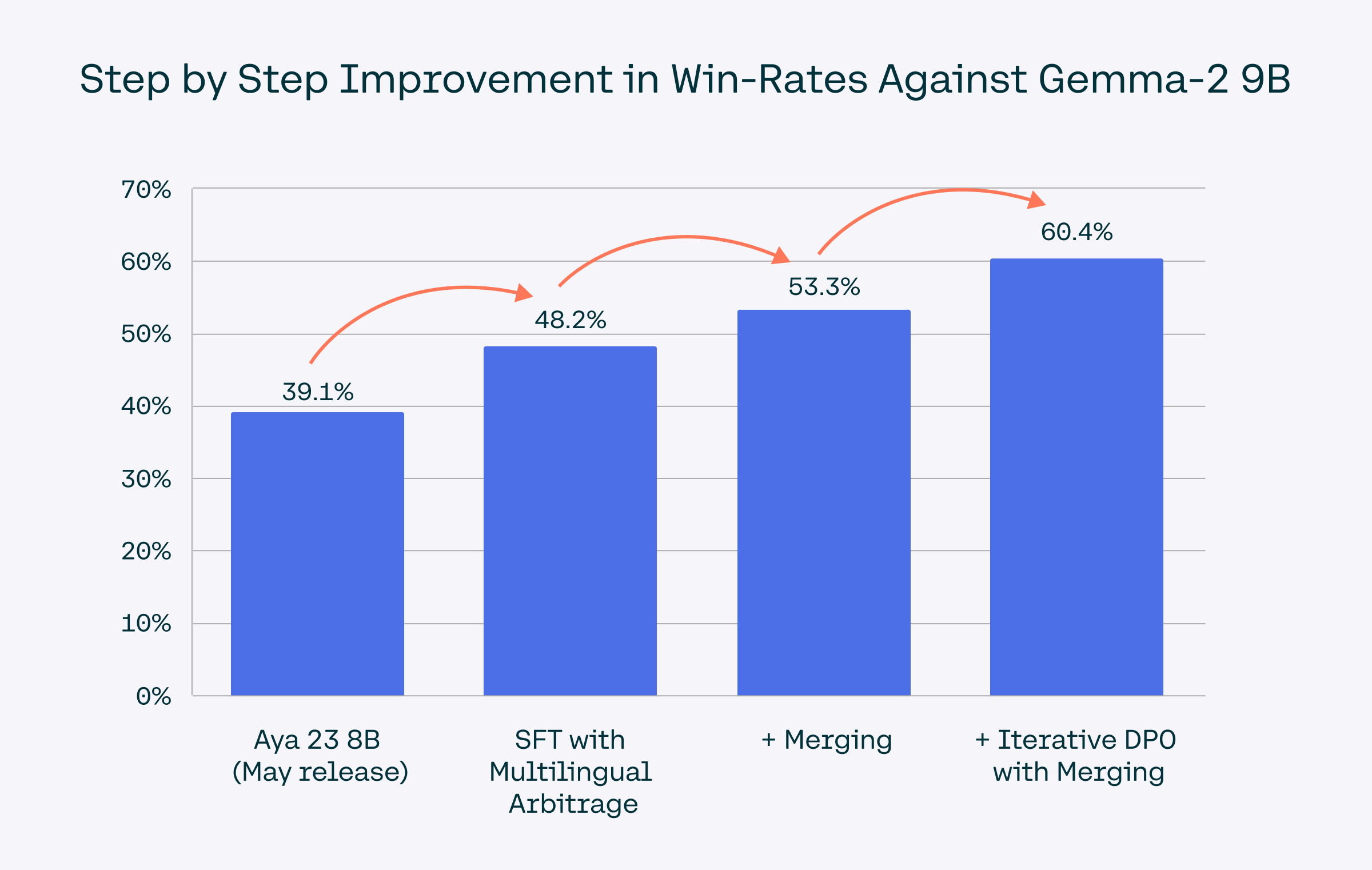}
      \end{subfigure}
      \hspace{0.1cm}
     \begin{subfigure}[b]{0.49\textwidth}
          \centering
          \includegraphics[width=\textwidth]{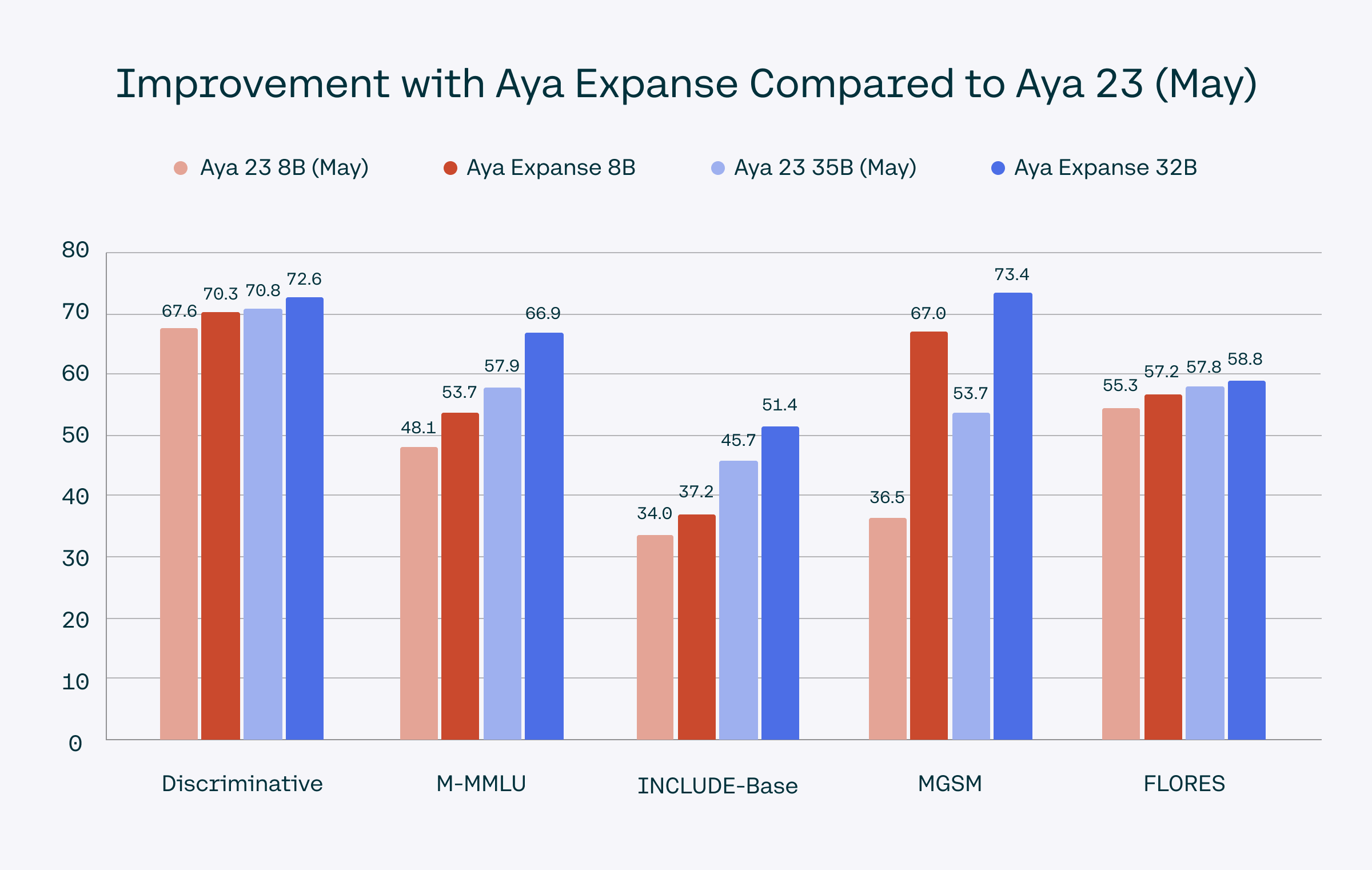}
      \end{subfigure}
      \caption{\textbf{Improvement with Aya Expanse models through Aya post-training recipe compared to their predecessor Aya 23 models}: On the left, we show win-rates against Gemma 2 9B on m-ArenaHard for each post-training step of Aya Expanse 8B and compared with Aya 23 8B \citep{aryabumi2024aya}. On the right, we compare Aya Expanse models with the previous Aya 23 model on academic benchmarks, showing significant improvement, especially for MGSM, Global-MMLU, and INCLUDE.}
      \label{fig:vs-aya-23}
 \end{figure}

\section{Results}

\subsection{Pairwise Model Comparisons for Open-ended Generations}

Figure \ref{fig:m-arenahard} and Figure \ref{fig:dolly} show pairwise win rates averaged across 23 languages between Aya Expanse models and the strong open-weight models, on m-ArenaHard and Dolly evaluation sets respectively. Overall, both Aya Expanse 8B and 32B models outperform all of the other models compared in their parameter class on both datasets.

\textbf{m-ArenaHard win rates.} On m-ArenaHard (Figure \ref{fig:m-arenahard}) where prompts are selected from hard and diverse instructions for high separability between models \citep{li2024crowdsourced}, Aya Expanse 8B outperforms other models up to 70.6\% win rate (vs Llama-3.1 8B). Among the models compared, Qwen-2.5 7B and Gemma-2 9B are the most competitive models where Aya Expanse 8B outperforms them by 55.7\% and 60.4\% win rates respectively. 

For the larger scale frontier models, Aya Expanse 32B achieves up to 76.6\% win rate (vs Mixtral 8x22B) on m-ArenaHard. Compared to the most competitive models, Gemma-2 27B and Qwen-2.5 32B, Aya Expanse achieves 51.8\% and 53.6\% win rates respectively. Remarkably, Aya Expanse 32B demonstrates superior performance against Llama-3.1 70B, achieving a 54.0\% win rate, even though Llama includes over twice as many parameters. 

\textbf{Dolly win rates.} On Dolly evaluation set \citep{ayadata2024}, we find that both 8B and 32B Aya Expanse models perform much better than other models compared to m-ArenaHard, where our 8B model achieves up to 83.9\% win rate (vs. Llama-3.1 8B) and 32B achieves up to 89.9\% win rate (vs Mixtral 8x22B). We relate the difference in win rates between m-ArenaHard and Dolly to the lesser degree of difficulty for evaluation prompts and the use of instructions from the same distribution (Dolly training set) in our training. 

\begin{wrapfigure}{R}{0.6\linewidth}
    \vspace{-0.5cm}
    \includegraphics[width=1.0\linewidth]{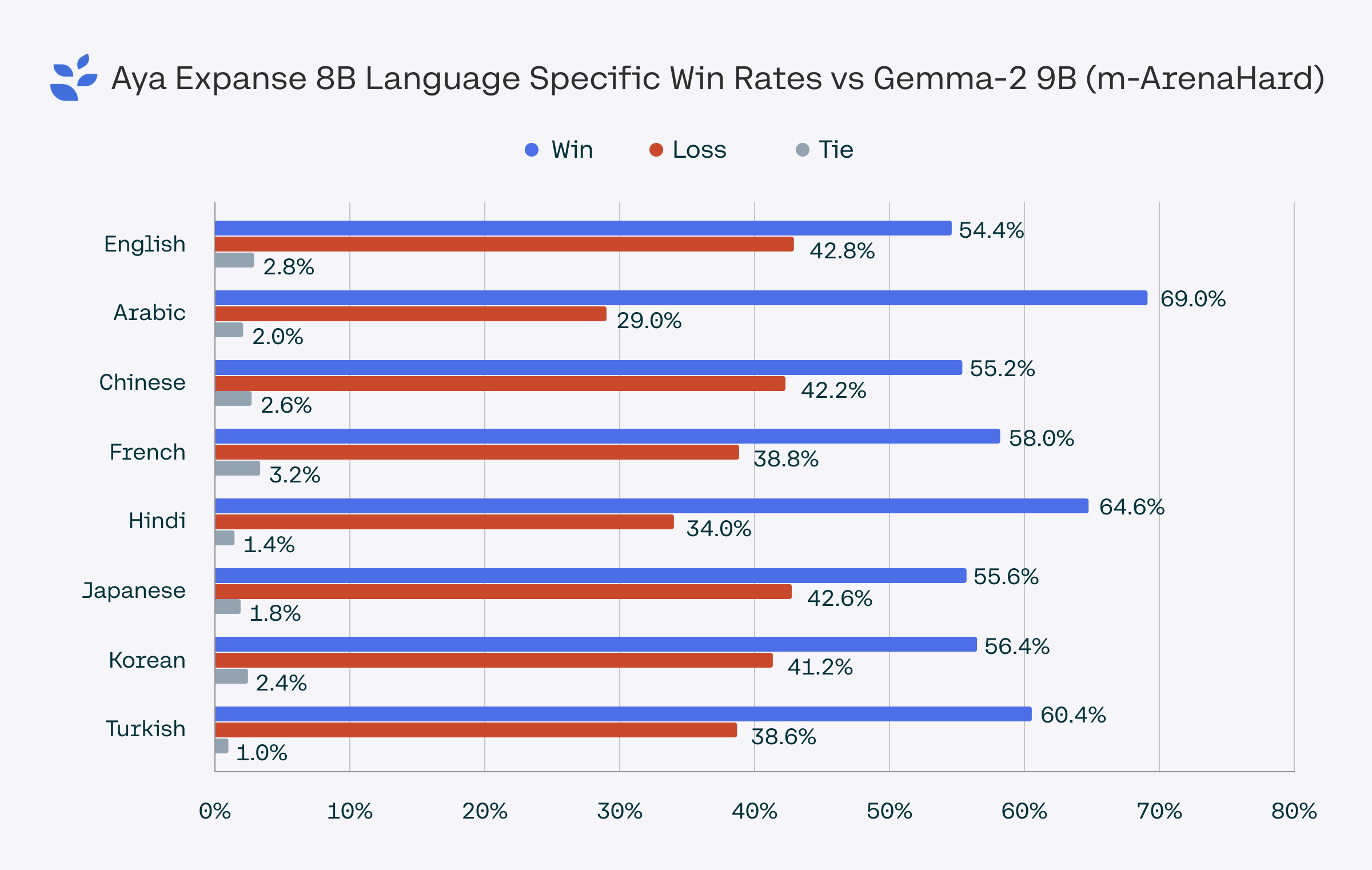}
    \caption{\textbf{Language-specific win-rates on m-ArenaHard}: Aya Expanse 8B performance against Gemma 2 9B for 8 diverse languages.}
    \label{fig:lang-specific-win-rates}
    \vspace{-0.5cm}
\end{wrapfigure}

\textbf{Consistent improvement across languages.}
While Aya Expanse outperforms the other open-weight models compared 
averaged across 23 languages, we find that Aya Expanse models also perform well individually in all 23 languages. As an example, in Figure \ref{fig:lang-specific-win-rates}, we show language-specific win-rates on m-ArenaHard for Aya Expanse 8B against Gemma 2 9B for selected 8 diverse languages. 
Aya Expanse 8B outperforms Gemma 2 9B  across all languages, including English, showing that it is possible to improve performance for many languages more equitably without sacrificing performance on high-resource languages like English.

\textbf{Step-by-step improvement with unified post-training recipe.} \label{sec:improve} As described in Section \ref{sec:post-training}, our post-training recipe consists of multilingual data arbitrage for supervised fine-tuning, \citep{odumakinde2024multilingualarbitrageoptimizingdata}, multilingual preference optimization \citep{dang2024rlhf}, and model merging \citep{aakanksha2024mix} in both SFT and each iteration of preference training. Figure \ref{fig:vs-aya-23} (left), we show the performance of Aya Expanse 8B on m-ArenaHard after each of these steps together with Aya 23 8B \citep[May release;][]{aryabumi2024aya}, compared to Gemma-2 9B as one of the most competitive model to ours. 

Aya Expanse 8B, even at the SFT stage, trained with multilingual arbitrage, had over 9.1\% (48.2\% vs 39.1\%) improvement in win rate measured against Gemma 2 9B compared to the previous Aya 23 model, demonstrating the effectiveness of this approach. Model merging at this stage enables a further 5.1\% improvement in win rate and finally, the combination of offline and online preference training leads to 7.1\% additional gains in win rate against Gemma 2 9B. Overall, with our unified post-training recipe, Aya Expanse 8B achieves a 60.4\% win rate against Gemma 2 9B where it shows a 20.3\% increase compared to the previous Aya 23 model. 


\textbf{Difference between GPT-4o and GPT-4o-mini.} \label{sec:gpt-4o-mini} We found that using GPT-4o-mini as a judge significantly biased win-rate results towards Aya Expanse by up to 10\%. For instance, Aya Expanse 8B achieves a win rate 9.3\% higher when using GPT-4o-mini as a judge against Gemma 2 9B, compared with GPT-4o as a judge. To avoid such biases, we report more conservative results where we use GPT-4o as a judge. 

\subsection{Multilingual Language Understanding}

\textbf{Discriminative benchmarks.} Table \ref{tab:8b-benchmarks} and \ref{tab:32b-benchmarks} include aggregated results for discriminative benchmarks for 8B and 32B model scales. We report the average score across XCOPA \citep{ponti2020xcopa}, XStoryCloze \citep{lin2021fewshot}, and XWinoGrad \citep{muennighoff2022crosslingual} where we compute 0-shot performance for all languages present for each dataset. 

At the 8B parameter class, while Gemma 2 9B performs best with 73.5 accuracy, Aya Expanse 8B achieves an accuracy of 70.3, competitive with Ministral 8B and outperforming Llama 3.1 8B and Qwen 2.5 7B. Notably, Aya Expanse 8B shows a 2.7 accuracy increase compared to Aya 23 8B \citep{aryabumi2024aya}. At the larger parameter scale, Aya Expanse 32B achieves an accuracy of 72.6, very competitive with Qwen 2.5 32B (73.4 Acc.) and Gemma 2 27B (72.9 Acc.) which includes a similar number of parameters. Llama 3.1 70B has the highest performance in this evaluation category, potentially leveraging a larger parameter count. Similar to the 8B scale, Aya Expanse 32B achieves a 1.8 accuracy increase compared to the previous Aya 23 35B model.   

\begin{table}
    \centering
    \resizebox{\textwidth}{!}{
\begin{NiceTabular}{@{}lccccc@{}}
\toprule
& \textsc{Discriminative} & \textsc{Global-MMLU} & \textsc{MGSM} & \multicolumn{2}{c}{\textsc{Translation (X$\leftrightarrow$E)}} \\
& (0-shot, Acc.) & (5-shot, Acc.) & (5-shot, Acc.) & (chrF\texttt{++}) & (xCOMET)\\
\midrule
Aya-23-8B (May) & 67.6 & 48.1 & 36.5 & 55.3 & 91.7 \\ 
\noalign{\smallskip} 
\hdashline 
\noalign{\smallskip} 
Llama-3.1-8B & 69.7 & 54.5 & 63.0 & 53.7 & 88.4 \\
Ministral-8B & 70.8 & 52.3 & 55.1 & 41.6 & 86.3 \\
Gemma-2-9B & \textbf{73.5} & 62.6 & 59.6 & 57.0 & 91.8 \\
Qwen-2.5-7B & 68.9 & \textbf{62.8} & 55.1 & 42.8 & 71.9 \\ 
\noalign{\smallskip} 
\hdashline 
\noalign{\smallskip} 
\includegraphics[scale=0.08]{./figures/logo2.png}Aya-Expanse-8B & 70.3 & 53.7 & \textbf{67.0} & \textbf{57.2} & \textbf{93.2} \\
\bottomrule
\end{NiceTabular}
}
\caption{\textbf{Academic multilingual benchmark results for 8-billion model class}: Evaluation setup includes the discriminative language understanding tasks (XWinograd, XCOPA, XStoryCloze), multilingual MMLU, MGSM, and machine translation. We compare Aya Expanse with Aya 23 (May release), Llama 3.1, Ministral, Gemma 2, and Qwen 2.5. All the competitor models are instruct fine-tuned (via SFT and RLHF).}
\label{tab:8b-benchmarks}
\end{table}

\textbf{Global-MMLU.} We report overall performance across 23 languages using both culturally agnostic and culturally sensitive subsets following \citep{singh2024globalmmluunderstandingaddressing}. Table \ref{tab:8b-benchmarks} and \ref{tab:32b-benchmarks} report results for the performance on 8B and 32B model scales. 

At the 8B scale, Qwen 2.5 7B and Gemma 2 9B show the highest result with 62.8 and 62.6 5-shot accuracies. Aya Expanse 8B achieves 53.7 accuracy, competitive with Llama 3.1 (54.5 Acc.) and outperforming Ministral 8B and previously released Aya 23 8B models. At the larger scale, Qwen 2.5 32B and Llama 3.1 70B share the top performance with 75.0 accuracy. Aya Expanse 32B with 66.9 accuracy performs close to Gemma 2 27B (68.3 Acc.) and outperforms both Mixtral 8x22B and Aya 23 35B. Similar to the other discriminative benchmarks, Aya Expanse models significantly improve multilingual MMLU scores or previous Aya 23 models: 5.6 and 9.0 increase in 5-shot accuracy for the 8B and 32B models respectively.

\textbf{INCLUDE.} To compare Aya Expanse models with their predecessor Aya 23 models, we use the INCLUDE-base benchmark \citep{romanou2024include} and use CoT prompting (\texttt{0-shot}) in the same language (\texttt{native_cot}). We report accuracy.
As shown in Figure \ref{fig:vs-aya-23} (right), Aya Expanse models show significant improvement over Aya 23 models. Concretely, Aya Expanse outperforms Aya 23 by 3.2 accuracy (37.2 vs 34.0) at 8B parameters and by 5.7 accuracy (51.4 vs 45.7) at 32B parameters scale.

\begin{table}
    \centering
    \resizebox{\textwidth}{!}{
\begin{NiceTabular}{@{}lcccccc@{}}
\toprule
& \textsc{Discriminative} & \textsc{Global-MMLU} & \textsc{MGSM} & \multicolumn{2}{c}{\textsc{Translation (X$\leftrightarrow$E)}} \\
& (0-shot, Acc.) & (5-shot, Acc.) & (5-shot, Acc.) & (chrF\texttt{++}) & (xCOMET)\\
\midrule
Aya-23-35B (May) & 70.8 & 57.9 & 53.7 & 57.8 & 93.1 \\ 
\noalign{\smallskip} 
\hdashline 
\noalign{\smallskip} 
Llama-3.1-70B & \textbf{75.3} & \textbf{75.0} & \textbf{78.0} & 54.2 & 88.3 \\
Mixtral-8x22B & 71.9 & 63.1 & 68.1 & 50.8 & 83.2 \\
Gemma-2-27B & 73.4 & 68.3 & 61.5 & 57.0 & 92.0 \\
Qwen-2.5-32B & 72.9 & \textbf{75.0} & 59.7 & 45.9 & 76.2 \\ 
\noalign{\smallskip} 
\hdashline 
\noalign{\smallskip} 
\includegraphics[scale=0.08]{./figures/logo2.png}Aya-Expanse-32B & 72.6 & 66.9 & 73.4 & \textbf{58.8} & \textbf{93.5} \\
\bottomrule
\end{NiceTabular}
}
\caption{\textbf{Academic multilingual benchmark results for 32-billion model class}: Evaluation setup includes the discriminative language understandign tasks (XWinograd, XCOPA, XStoryCloze), multilingual MMLU, MGSM and machine translation. We compare Aya Expanse with Aya 23 (May release), Llama 3.1 , Mixtral, Gemma 2 and Qwen 2.5. All the competitor models are instruct fine-tuned (via SFT and RLHF). Note that both Mixtral (141B total, 39B active) and Llama 3.1 (70B) includes over twice as many parameters than Aya Expanse models in this comparison. 
}
\label{tab:32b-benchmarks}
\end{table}

\subsection{Multilingual Mathematical Reasoning}
Multilingual Grade School Math Benchmark (MGSM) is the academic benchmark that the Aya Expanse models achieve a large improvement compared to their predecessor Aya 23 models. As shown in Table \ref{tab:8b-benchmarks} and \ref{tab:32b-benchmarks}, Aya Expanse 8B achieves over twice the accuracy compared to Aya 23 8B (36.6 vs 67.0) and Aya Expanse 32B shows 19.7 accuracy increase over Aya 23 35B. These significant improvements are the results of our improved post-training recipe.

Compared to the other models, the Aya Expanse model performs best in their parameter class in this task. Aya Expanse 8B with 67.0 accuracy outperforms all other models in the 8B parameter class. Similarly, Aya Expanse 32B with 73.4 accuracy outperforms Qwen 2.5 32B (59.7 Acc.) and Gemma 2 27B (61.5 Acc.). From the models compared, only Llama 3.1 70B performs better than Aya Expanse 32B (78.0 vs 73.4), however, it has over twice as many parameters. 

\subsection{Machine Translation}
As the final category, we evaluate the machine translation performance of Aya Expanse models and the other models on FLORES \citep{goyal-etal-2022-flores,nllb2022}. Table \ref{tab:8b-benchmarks} and \ref{tab:32b-benchmarks} report average chrF\texttt{++} and xCOMET for 22 language pairs in both X$\rightarrow$English and English$\rightarrow$X directions. In this category, Aya Expanse models achieve the best performance in both 8B and 32B parameter classes. 

At the 8B scale, Aya Expanse 8B outperforms the second-best model Gemma 2 9B by 0.2 chrF\texttt{++} (57.2 vs 57.0) and 1.4 xCOMET score (93.2 vs 91.8) a strong improvement equivalent to an effect size of +2.9 BLEU, as shown by \cite{kocmi-etal-2024-navigating}. 
Notably after these two models, Aya 23 8B still performs as the third-best model significantly outperforming Llama 3.1 8B, Qwen 2.5 7B, and Ministral 8B. 

At the 32B parameter scale, Aya Expanse 32B shows the top performance with 58.8 chrF\texttt{++} and 93.5 xCOMET scores while Aya 23 35B achieves the second-best performance (57.8 chrF\texttt{++}, 93.1 xCOMET). Gemma 2 27 follows these two models with competitive performance, however, the other models fall significantly behind, including the ones that have many more parameters such as Llama 3.1 70B and Mixtral 8x22B.    

\section{Conclusion}

Although there have been significant advancements in language technologies recently, this progress remains primarily focused on English. To address this gap, we present the Aya Expanse model family, which is developed through a suite of innovative methodologies, including multilingual data arbitrage~\citep{odumakinde2024multilingualarbitrageoptimizingdata} which strategically samples from a diverse pool of teacher models to generate high-quality synthetic multilingual datasets, multilingual preference optimization and safety tuning~\citep{dang2024rlhf,aakanksha2024multilingual} which enhance model alignment with human preferences across languages, and model merging~\citep{aakanksha2024mix} that optimize performance by leveraging cross-lingual transfer and diversity among language families. These innovations have empowered us to create models that not only excel in multiple languages but also sustain competitive performance across a diverse array of tasks. Our evaluations show that Aya expanse models as a result of these technological innovations, establish a new frontier in multilingual performance. By making these model weights publicly available, we aim to contribute to future research and further this crucial mission.

\section{Acknowledgements}

We thank the Hugging Face team for helping us with our open-weights release including Omar Sanseviero, Pedro Cuenca, Vaibhav Srivastav, Lysandre Debut, Aritra Roy Gosthipaty.

Thanks to colleagues who have supported various aspects of this project: Aidan Peppin, Arielle Bailey, Sarah Elsharkawy, Julia Kreutzer, Kosta Starostin, Sitong Sun, Aakanksha, Seraphina Goldfarb-Tarrant, Kelly Marchisio, Walter Beller-Morales, Anubhav Sachan, Saurabh Dash, Sebastian Ruder, Manuel Nader, Bolaji Agunbiade, Marwan Ahmed, Morgan Normal, Jerald Aguilar, Josh Gartner, Kyle Lastovica, Wojciech Kryscinski, Johnny Nguyen, Kate Svetlakova, Adam Sinton, Ana Cismaru, Lisa Alazraki, Jay Alamar, Sandrine Vaillancourt, Isabelle Camp. 
We thank Cohere and Cohere For AI teams for support across multiple Aya releases. 
Finally, we thank Cohere For AI Community for testing the models throughout the model release. 

\bibliography{main,addon}

\end{document}